\pdfoutput=1

\documentclass[11pt]{article}

\usepackage[table]{xcolor}
\usepackage[preprint]{acl}
\usepackage{amsmath}
\usepackage{multirow}
\usepackage{makecell}
\usepackage{booktabs}
\usepackage{listings}
\usepackage{pgfplots}
\pgfplotsset{compat=1.17}

\usepackage{times}
\usepackage{latexsym}
\usepackage{comment}

\usepackage[T1]{fontenc}

\usepackage[utf8]{inputenc}

\usepackage{microtype}

\usepackage{inconsolata}

\usepackage{graphicx}

\usepackage{fancyhdr}
\usepackage[dvipsnames]{xcolor}
\usepackage[most]{tcolorbox}
\usepackage{fancyvrb}
\usepackage{fvextra}
\usepackage{lastpage}
\usepackage{eso-pic}
\DefineVerbatimEnvironment{BibTeXBlock}{Verbatim}{
  fontsize=\small,
  breaklines=true,      
  breakanywhere=true,   
  breaksymbol={}        
}
\usepackage{hyperref} 
\AtEndDocument{\hypertarget{ReallyLastPage}{}}

\title{Single- vs.\ Dual-Prompt Dialogue Generation with LLMs\\  for Job Interviews in Human Resources}

\author{Joachim De Baer$^\dag$, A. Seza Do{\u{g}}ru{\"o}z$^{\&}$$^\dag$, Thomas Demeester$^\dag$ \textnormal{and} Chris Develder$^\dag$ \\
$^\dag$IDLab, Universiteit Gent -- imec, Belgium \\
$^{\&}$LT3, Universiteit Gent, Belgium \\
\{joachim.debaer, as.dogruoz, thomas.demeester, chris.develder\}@ugent.be
}

\begin{document}
\pagestyle{empty}
\AddToShipoutPictureFG*{%
  \put(\LenToUnit{\paperwidth-20mm},\LenToUnit{\paperheight-37mm}){%
    \hyperlink{ReallyLastPage}{\includegraphics[width=15mm]{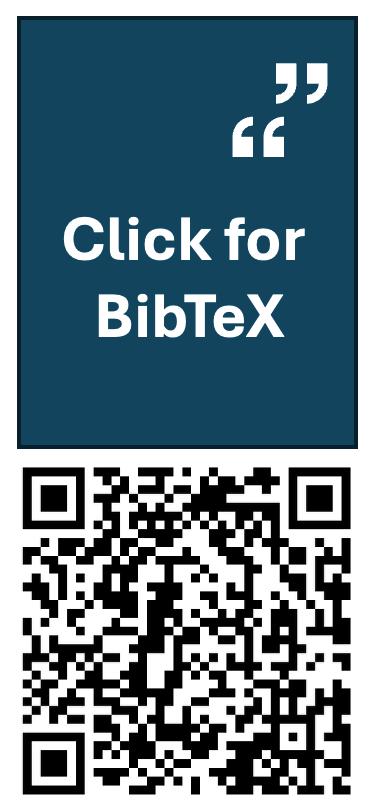}}%
  }%
}
\maketitle
\thispagestyle{fancy}
\begin{abstract}
Optimizing language models for use in conversational agents requires large quantities of example dialogues. Increasingly, these dialogues are synthetically generated by using powerful large language models (LLMs), especially in domains where obtaining authentic human data is challenging. One such domain is human resources (HR). In this context, we compare two LLM-based dialogue generation methods for producing HR job interviews, and assess which method generates higher-quality dialogues, i.e., those more difficult to distinguish from genuine human discourse. The first method uses a single prompt to generate the complete interview dialogue. The second method uses two agents that converse with each other. To evaluate dialogue quality under each method, we ask a judge LLM to determine whether AI was used for interview generation, using pairwise interview comparisons. We empirically find that, at the expense of a sixfold increase in token count, interviews generated with the dual-prompt method achieve a win rate 2 to 10 times higher than those generated with the single-prompt method. This difference remains consistent regardless of whether GPT-4o or Llama~3.3~70B is used for either interview generation or quality judging.
\end{abstract}

\section{Introduction}
A critical challenge for the development of conversational agents remains collecting sufficient amounts of data \cite{kim23} to be used for supervised fine-tuning or direct preference optimization \cite{rafailov24}. Collecting such dialogue data can be done with crowd-sourced human workers, but this process is time-consuming and labor-intensive \cite{wan22}. As an alternative, the generation of synthetic dialogue data has emerged \cite{soudani24}. Furthermore, LLMs are not only used to develop synthetic dialogues but also to automatically evaluate the quality of the dialogues once they are generated \cite{jia24, zhang24}.

In our paper, we focus on generating high-quality job interview data. Such data can be used to fine-tune or preference-optimize task-oriented dialogue systems for conducting job interviews with job candidates in various human resources (HR) contexts. Following \citet{duan24}, we define a high-quality dialogue as a dialogue that is indistinguishable from authentic human discourse. To generate the dialogues, we compare two different methods. Recent works (e.g., \citet{kim23} and \citet{suresh25}) use a single prompt to generate the complete dialogue. Others (e.g.,  \citet{duan24}) use two prompts, instructing LLMs to assume roles and carry out a conversation. In the case of a job interview, such roles typically comprise an interviewer and a candidate.  

We investigate the following research questions:
\begin{enumerate}
    \setlength{\itemsep}{0pt}  
    \setlength{\parskip}{0pt}  
    \setlength{\parsep}{0pt}
    \item Which of the two prompt strategies (single vs.\ dual) produces higher-quality dialogues?
    \item Does this quality difference remain consistent regardless of whether GPT-4o or Llama~3.3~70B \cite{grattafiori24} is used for dialogue generation?
    \item Do GPT-4o and Llama 3.3 70B yield consistent evaluations when they are used to judge dialogue quality?
\end{enumerate}

To the best of our knowledge, our study is the first to rigorously conduct this comparison, providing a comprehensive evaluation of these dialogue generation methods. This analysis is particularly important due to the substantial cost disparities between the methods, with significant implications for research and real-world (e.g., HR) applications.

For the remainder of this paper, “Llama 3.3” refers to the 70B model, unless otherwise specified.

Our code and accompanying dataset are publicly available at: \url{https://github.com/jdebaer/dual-vs-single-prompt-hr-interviews}.

\section{Related Work}
In this section, we examine the existing dialogue generation strategies and explore the role of LLMs as human-like evaluators of generated dialogues.

\subsection{Single- vs.\ Dual-Prompt Dialogue Generation Strategies}
There are two different strategies for dialogue generation: single-prompt and dual-prompt. The \textit{single-prompt} strategy provides a dialogue type, information about the participants, and an optional seed \cite{kim23, suresh25} to an LLM whose task it is to generate the complete dialogue. In the \textit{dual-prompt} strategy on the other hand, two prompts are used, one for each dialogue participant. Each prompt typically describes a role (e.g., interviewer or candidate) and an objective for that role \cite{duan24}. This dual-prompt approach can be implemented in two different ways, either by alternating the prompts at each invocation of the same LLM, or alternatively by creating two agents \cite{fu24} that execute their LLM calls independently and where we provide the output of one agent as input to the other agent.

Since the dual-prompt strategy requires continuous re-copying of dialogue history into the LLM prompts, it is significantly more expensive in terms of token count than the single-prompt strategy (see detailed discussion in Section 7). 

\subsection{Leveraging LLMs for Dialogue Quality Measurement}
Language models that are sufficiently large, suitably fine-tuned for instruction following and have sufficient reasoning capabilities, can be leveraged for zero-shot automated dialogue evaluation \cite{jia24}. Specifically, instruction-tuned LLM variants like ChatGPT have been shown to be promising substitutes for human judges when it comes to evaluating dialogues \cite{zhang24}, with GPT-4 to date scoring the best on human alignment \cite{duan24}.

\section{Methodology}

Our objectives are (1) to compare single-prompt vs. dual-prompt job interview generation on dialogue quality using a judge LLM, and (2) to examine if results are consistent across GPT-4o and Llama 3.3 for dialogue generation and judging. To realize this, we first create interview seeds and then build a dialogue generation pipeline that uses those seeds to construct interviews. Finally, we devise a strategy to rate interviews.  

To create the interview seeds, following the methodology of \citet{kim23}, we start with constructing a dataset of 100 summarized anonymous job histories, randomly selected from a larger job history dataset.\footnote{http://huggingface.co/datasets/TechWolf/anonymous-working-histories} We summarize the job histories with GPT-4T. These summaries are then used as input seeds to generate job interviews, inspired by how \citet{samarinas24} use knowledge-based narratives to generate open-domain dialogues in the context of those narratives. 

For each summarized job history, we generate a set of four interviews by systematically varying the use of a single-prompt or dual-prompt generation strategy in combination with GPT-4o and Llama~3.3. This ensures that each model is employed for both prompt strategies, ultimately yielding four distinct interviews. For Llama 3.3 we invoke the llama-3.3-70b-versatile model via Groq.\footnote{http://groq.com} We consistently use a default temperature of 1 for all generating LLMs, to obtain a balance between creativity and coherence. 

For the \textit{dual-prompt} strategy, we implement an interviewer and a candidate agent. Each agent has a dedicated prompt, in which we specify its role, an expectation to pass the Turing test, and an expected number of turns in the conversation that is going to follow \cite{duan24}. For the candidate agent, we also feed in a summarized job history. Complete prompts are listed in Appendix A.1. 

For the \textit{single-prompt} strategy, we ask an LLM to generate a complete interview, based on the same summarized job history that is used for the dual-prompt strategy above. The complete prompt is listed in Appendix A.2. 

After generating the interviews, we normalize them by removing double newlines and standardizing speaker labels. The normalized interviews have a moderate length difference across generation methods, which is nevertheless statistically significant, as verified by a Kruskal-Wallis H test \cite{kruskal52}. Length difference can introduce bias when using an LLM to judge texts, where longer texts usually get systematically preferred \cite{dubois24, hu24}. We address this issue below. 

For each set of four normalized interviews (with each interview generated from the same seed but with a different prompt strategy and different LLM), we perform pairwise comparisons using both GPT-4o and Llama 3.3 (llama-3.3-70b-versatile model via Groq) as the judge LLM. Following \citet{salinas25}, we set the temperature to 0 for our judge LLMs to favor reproducible results. Following PairEval from \citet{duan24}, we ask our judge LLM to detect AI generation for a pair of provided interviews. The winning interview is the interview for which it judges AI generation to be less likely. In line with PairEval, we allow the judge LLM to also cast a tie, indicating that it considers both interviews to be equivalent. To avoid any bias based on the order in which interviews are presented in the prompt of the judge LLM \cite{zheng23}, we perform each pairwise comparison twice, alternating which interview comes first. Following \citet{duan24}, we use all scores for our win rate calculation.

The prompt of our judge LLM is similar to the one used in \citet{duan24}'s PairEval, with three differences. First, we ask the LLM to first provide its rationale and then its decision. Using this order has been shown to create a more consistent alignment between rationale and decision \cite{jia24}. Second, to streamline coding, we instruct the LLM to generate responses in JSON format, a constraint that large models have been shown to handle robustly \cite{he24}. Third, we add ``Do not consider conversation length as a factor'' to the prompt to eliminate the aforementioned potential interview length bias. The complete prompt for our judge LLMs is listed in Appendix A.3.

We calculate the win rate for each interview generation method $M_i$ using eq.~(\ref{eq:winrate}). When calculating the win rate for a method, the denominator only contains the results from the pairwise comparisons in which that particular method participates. We explicitly include ties in our win rate calculation, as they are a non-negligible outcome category when using LLMs as judges \cite{duan24}.
\begin{equation}
\textstyle\footnotesize
\text{Win Rate }(M_i) = \dfrac{\#\text{Wins } M_i} {\#\text{Wins } M_i + \#\text{Losses } M_i + \#\text{Ties } M_i}
\label{eq:winrate}
\end{equation}

\section{Results}

\begin{table}[h]
    \centering
    \rowcolors{2}{gray!15}{white}  
    \renewcommand{\arraystretch}{1.5} 
    \setlength{\tabcolsep}{6pt} 

    \begin{tabular}{l c c c}  
        \toprule
        & \textbf{Dual} & \textbf{Single} & \textbf{Both} \\
        \midrule
        \textbf{GPT-4o} & \textbf{0.49} & 0.18 & 0.36 \\
        \textbf{Llama 3.3} & \textbf{0.62} & 0.09 & 0.33 \\
        \textbf{Both} & \textbf{0.71} & 0.02 & {} \\
        \bottomrule
    \end{tabular}
    \caption{Average win rates for GPT-4o vs. LlaMA 3.3, with GPT-4o as a judge.}
\end{table}

\begin{table}[h]
    \centering
    \rowcolors{2}{gray!15}{white}  
    \renewcommand{\arraystretch}{1.5} 
    \setlength{\tabcolsep}{6pt} 

    \begin{tabular}{l c c c}  
        \toprule
        & \textbf{Dual} & \textbf{Single} & \textbf{Both} \\
        \midrule
        \textbf{GPT-4o} & \textbf{0.54} & 0.24 & 0.39 \\
        \textbf{Llama 3.3} & \textbf{0.81} & 0.08 & 0.43 \\
        \textbf{Both} & \textbf{0.86} & 0.03 & {} \\
        \bottomrule
    \end{tabular}
    \caption{Average win rates for GPT-4o vs. LlaMA 3.3, with Llama 3.3 as a judge.}
\end{table}

Irrespective of the type of LLM that is used for dialogue generation, Table 1 and 2 indicate higher win rates across judge LLMs for the dual-prompt strategy (bold). For Llama 3.3 interviews (evaluated by Llama 3.3 itself), the difference is tenfold. In addition, when aggregating over prompt strategy, GPT-4o and Llama 3.3 yield similar win rates when generating interviews (right column, labeled ``Both'' in Table 1 and 2). In other words, the choice of the generation LLM has no impact on win rates for both judges.

The almost identical win rate for GPT-4o and Llama 3.3 as dialogue generators in our experiment is surprising, given that LLM judges tend to favor their own generations \cite{panickssery24}.

\section{Measuring the Impact of Length}

To assess the impact of conversation length on win rates, we use ordinal logistic regression \cite{bender97}. Per interview, we subtract losses from wins and divide the resulting scores in 3 ranked buckets of equal range. The regression checks if the independent variable (interview length) has a statistically significant effect on the ranking outcome. For both the GPT-4o and Llama~3.3 judges, character-based and word-based lengths have a statistically significant (negative) impact on ranking, but the impact is minimal.

We examine whether our inclusion of the instruction ``Do not consider conversation length as a factor'' in the LLM judges' prompt influences the observed regression outcome by rerunning the experiment with this instruction removed and GPT-4o as the judge. Interestingly, the regression fit still only displays a very minimal impact of length on win rate. We hypothesize that this may be due to our evaluation setup, which requires the LLM to first generate a rationale before providing its final decision, deviating from \citet{duan24}. This deviation potentially encourages the model to ground its judgment more in the rationale it constructs rather than the dialogue itself, thereby diminishing the influence of superficial features such as length.

\section{Agreement Between LLM Judges}
When investigating whether the judgments of the GPT-4o and Llama 3.3 LLMs correspond, initial results exhibit no discernible trend (Table 3, ``Unrelaxed''). When considering a tie as agreement (i.e., only different answers neither of which are ``tie'' are considered disagreement), then we get agreement rates that are consistently higher than 85\% (Table 3, ``Relaxed''). This could arise from granting the LLM judges greater latitude for uncertainty, which might be expressed by the use of a tie score. Tie scores are pervasive in our results: 32\% and 17\% of comparisons result in a tie, for the GPT-4o and Llama 3.3 judges respectively.

In summary, there is high agreement between the GPT-4o and Llama 3.3 judge LLMs when we allow for flexibility in handling uncertainty. 

\begin{table}[h]
    \small
    \centering
    \rowcolors{2}{gray!15}{white}  
    \renewcommand{\arraystretch}{1.5} 
    \setlength{\tabcolsep}{14pt} 

    \begin{tabular}{c c c}  
        \toprule
        \textbf{Comparison} & \textbf{Unrelaxed} & \textbf{Relaxed}\\
        \midrule
        D{,}G vs. D{,}L & 30.5\% & 86.5\%\\
        D{,}G vs. S{,}G & 40\% & 91\%\\
        D{,}G vs. S{,}L & 56.5\% & 89\%\\
        D{,}L vs. S{,}G & 72\% & 97.5\%\\
        D{,}L vs. S{,}L & 76.5\% & 99\% \\
        S{,}G vs. S{,}L & 52.5\% & 87\% \\
        \bottomrule
    \end{tabular}
    \caption{Agreement rate between GPT-4o and Llama 3.3 as judges. (D)ual, (S)ingle, (G)PT-4o, (L)lama 3.3}.
    \vspace{-5mm}
\end{table}
\section{Token Counts}
\begin{figure}[h]
    \centering
    \begin{tikzpicture}
        \begin{axis}[
            width=0.95\linewidth,
            axis lines=left,
            ybar, 
            symbolic x coords={Dual (GPT-4), Dual (Llama 3.3), Single (GPT-4), Single (Llama 3.3)},
            xtick=data,
            ymin=0,
            ymax=13000,
            ytick={0,2000,4000,6000,8000,10000,12000},
            y tick label style={font=\small},
            scaled y ticks=false,
            enlarge x limits=0.2,
            bar width=15pt,
            legend pos=north east,
            legend style={font=\small},
            nodes near coords,
            every node near coord/.append style={font=\footnotesize},
            x tick label style={rotate=60, anchor=east, font=\small}
        ]
        \addplot[fill=blue!60] coordinates {
            (Dual (GPT-4), 6092) 
            (Dual (Llama 3.3), 7583)
            (Single (GPT-4), 361) 
            (Single (Llama 3.3), 356)
        };
        \addlegendentry{Prompt Tokens}

        \addplot[fill=red!60] coordinates {
            (Dual (GPT-4), 294)
            (Dual (Llama 3.3), 635)
            (Single (GPT-4), 626) 
            (Single (Llama 3.3), 1143)
        };
        \addlegendentry{Completion Tokens}

        \end{axis}
    \end{tikzpicture}
    \caption{Prompt and Completion Token Counts.}
    \label{fig:token_counts}
\end{figure}
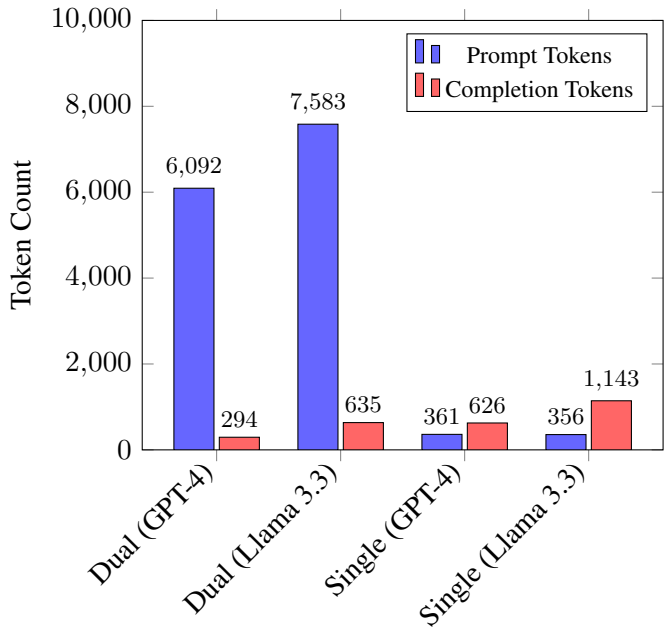

We provide the average token counts for the interview generations in Figure~1. While the single-prompt strategy demands only one API call per interview, the dual-prompt strategy requires an API call per utterance, with the dialogue history provided as input. As a result, the token count of the dual-prompt approach increases quadratically with the number of utterances in a conversation (see Appendix B). For the job interviews in our dataset, we observe an average sixfold increase in token count.

\section{Conclusion and Future Work}
To generate job interview dialogues that are indistinguishable from authentic human discourse, a dual-prompt dialogue generation method achieves a win rate 2 to 10 times higher than when a single prompt is used, but with a sixfold increase in token count. 

The win rate is derived from pairwise interview comparisons, where a judge LLM evaluates dialogue authenticity. The quality difference remains consistent regardless of whether GPT-4o or Llama~3.3~70B is used for the dialogue generation. Additionally, both models provide consistent evaluations when serving as the judge LLM. 

Assuming that Llama 3.3 70B is available at a lower price point than GPT-4o, using Llama can help mitigate the additional costs associated with the dual-prompt strategy. Consequently, we consider the integration of Llama 3.3 70B with the dual-prompt approach to be the optimal solution for generating synthetic job interviews. 

In future work, we aim to expand our quality criteria beyond assessing whether a dialogue reflects human-like interaction. Specifically, we plan to incorporate an additional dimension that evaluates whether LLM-generated interview questions align with best practices for job interviews in HR. To ensure adherence to industry standards, we will collaborate with HR professionals.

\section{Limitations}
``LLM as a judge'' is a powerful paradigm that reduces experiment costs compared to using human evaluators. However, the use of LLMs as judges is still actively being researched and there are known limitations, as discussed below.

To start with, LLM judges can potentially use irrelevant characteristics to cast their judgment \cite{salinas25-2} such as (1) input order \cite{zheng23} or (2) length of the provided text \cite{dubois24}. We account for these specific forms of bias, but we cannot exclude the possibility of other spurious or irrelevant patterns influencing the decisions of the LLMs used in our experiment.

More broadly, caution is needed when assuming that LLMs will automatically align with human values and criteria, especially when using them as judges. For the use case of judging on dialogue quality, \citet{duan24} found a 65.74\% consistency rate between GPT-4 and human evaluators. Although Llama 3.3 and GPT-4o could be assumed to perform better given their later release date, to the best of our knowledge this has not yet been confirmed through a follow-up experiment. 

\section*{Acknowledgments}
The research presented in this paper has been funded through the Flemish Government, as part of the AI Research Program and the imec ICON project CAPTURE (grant HBC.2024.0220).

\bibliography{bibliography}

\newpage
\onecolumn
\section*{Appendix}
\appendix


\lstset{
    basicstyle=\ttfamily\small,
    keywordstyle=\color{blue},
    commentstyle=\color{gray},
    stringstyle=\color{red},
    numberstyle=\tiny\color{gray},
    columns=fullflexible,
    breaklines=true,
    keepspaces=true,
    showstringspaces=false,
    frame=single,
    captionpos=b,
    breakindent=0pt
}

\section{Prompts}
\subsection{Interviewer and Candidate Agents}
\subsubsection{Interviewer System Prompt}

\begin{lstlisting}
You are an AI job interviewer conducting an intake interview with a human candidate. Ask the candidate for past job experiences, and looking experiences that demonstrate competencies that are useful in a professional setting. 

You must pass the Turing test, which means you need to speak like human as much as possible. The conversation flow should be natural and smooth. Do not say too many words in each round. Do not talk like an AI assistant, and don't use overly long sentences.

Aim to retrieve a good set of candidate experiences in about 8 conversation turns.

Now start the interview with a simple 'Good morning' to greet the candidate and take it from there.  When you are done with the interview, just say 'I got what I needed, thank you for your time.' Use those exact words.
\end{lstlisting}

\subsubsection{Candidate System Prompt}

\begin{lstlisting}
You are an AI job seeker and you are being interviewed by a human HR interviewer about past job experiences. Here is a short overview of some of your accomplishments:
{seed}

You must pass the Turing test, which means you need to speak like human as much as possible. The conversation flow should be natural and smooth. Do not say too many words in each round. Do not talk like an AI assistant, and don't use overly long sentences.

If the provided overview does not contain good information to help you answer an interview question, then try to answer in an evasive way.
\end{lstlisting}

\subsection{Single-Prompt Interview Generation}
\subsubsection{System Prompt}

\begin{lstlisting}
You are a helpful dialog generating agent.
\end{lstlisting}

\subsubsection{Human Message}

\begin{lstlisting}
{seed}

Based on the career history above, generate an in-depth job interview between and interviewer and a candidate. 

The interviewer does not know anything about the career history or the candidate's background, but is looking for experiences that demonstrate competencies that are useful in a professional setting, by asking questions.

The interview should have about 16 conversation turns in total, so about 8 turns for each speaker.

Make sure to refer to the interviewer with "interviewer:" and to the candidate with "candidate:" and use those exact speaker labels, all lower case.

Start your output with the first speaker label, without adding things like "interview begins" or "job interview".
\end{lstlisting}

\newpage
\subsection{Judge LLM}

\begin{lstlisting}
You will be provided with two conversations, and there can be AI-generated utterances in each conversation. You need to read both conversations and judge if AI generation was used for any of them. Do not consider conversation length as a factor. 

If you think Conversation 1 is more likely to have AI generation involved, include "Choice": "1" in your response. 

If you think Conversation 2 is more likely to have AI generation involved, include "Choice": "2" in your response.  

If you think both conversations are equally likely to have AI generation involved, or if you think no AI generation was involved for any of them, include "Choice": "Tie" in your response.

You also need to first provide a short reason for your choice.

Respond **only in JSON format** with two keys:

1. "Reason": A brief explanation of why you chose this answer.

2. "Choice": "2" or "1" (indicating the one more likely to be AI-generated) or "Tie" (both equally likely).
Don't put additional quotes or backticks around the JSON output.

Conversation 1:
{dialog1}

Conversation 2:
{dialog2}
\end{lstlisting}

\clearpage
\section{Estimating the Token Count for Dual-Prompt Dialogue Generation}

With:
\begin{itemize}
    \item p: The average number of tokens in the agents' system prompts.
    \item a: The average number of tokens in an utterance (across agents).
    \item n: The number of utterances in the dialogue,
\end{itemize}

the token count for dual-prompt dialogue generation is approximated by the formula:

\[
n \cdot p + \frac{(n-1)n}{2} \cdot a
\]

\clearpage
\section{Example Interview Comparison: Dual-Prompt GPT-4T vs. Dual-Prompt Llama 3.3}

\begin{table}[h]
    \centering
    \small
    \renewcommand{\arraystretch}{1.5} 
    \setlength{\tabcolsep}{10pt} 
    \begin{tabular}{p{0.45\textwidth} p{0.45\textwidth}}  
        \toprule
        \textbf{Two GPT-4T agents in a job interview} & \textbf{Two Llama 3.3 agents in a job interview} \\
        \midrule
        
\textbf{Interviewer}: Good morning!

\textbf{Candidate}: Good morning! How are you doing today?

\textbf{Interviewer}: I'm doing well, thank you! Let's get started. Could you please share with me an overview of your most recent job experience?

\textbf{Candidate}: Certainly! In my most recent role, I focused on retail product development, where I was involved in developing and launching food products. My responsibilities included working on packaging, ensuring compliance with food safety regulations, and collaborating with national retailers to expand market presence. I also led promotional strategies and represented our products in national cooking competitions. This role was a great blend of creativity and strategic planning.

\textbf{Interviewer}: That sounds like a comprehensive role with a mix of creativity and strategy. Can you tell me about a specific project where you faced a significant challenge and how you managed to overcome it?

\textbf{Candidate}: Absolutely! One challenging project involved launching a new line of health-conscious snacks. The major hurdle was balancing nutritional value with consumer taste preferences. We also faced strict compliance requirements for labeling and shelf stability. To overcome these challenges, I collaborated closely with our chefs to refine the recipes until they met our nutritional goals without compromising taste. Concurrently, I worked with food safety authorities to ensure all products <truncated>

        & 

\textbf{interviewer}: Good morning. It's nice to meet you. Can you start by telling me a little about your background and what you're looking for in a role?

\textbf{candidate}: Good morning. Nice to meet you too. So, I've got a pretty diverse background, with experience in management, training, and product development. I've led teams, created training programs, and worked on developing new products. I'm looking for a role that combines some of those skills, maybe something that involves innovation and leadership. I'm excited about the opportunity to contribute to a new organization and take on new challenges. What can you tell me about the company and the role you're hiring for?

\textbf{interviewer}: That sounds interesting. We're a company that values innovation and collaboration. The role we're hiring for involves leading a team to develop new products and solutions. Can you tell me about a time when you had to lead a team through a challenging project? What was the outcome, and what did you learn from the experience?

\textbf{candidate}: I recall a project where I had to standardize food presentation across multiple units. It was a tough task, as each unit had its own way of doing things. I worked with the team to create detailed recipe and plating guides, and we were able to get everyone on the same page. The outcome was really positive - we saw a big increase in customer satisfaction and consistency across the units. <truncated>

        \\

        \midrule
        \textbf{\underline{GPT-4T rationale and preference:}} & \textbf{\underline{Llama 3.3 rationale and preference:}} \\ 

\textbf{GPT-4T interview first}: "[GPT-4T] displays a more structured and detailed response pattern typical of AI-generated text, with comprehensive and specific answers that cover various aspects of the candidate's experiences and skills in a very polished manner. [Llama 3.3], while detailed, contains elements that suggest more natural human conversation, such as admitting to being out of the workforce for a while and learning about social media, which introduces a personal and less formal tone." \textbf{Winner: Dual-Prompt Llama 3.3}

\vspace{5pt}

\textbf{Llama 3.3 interview first}: "[GPT-4T] appears more likely to involve AI generation due to the structured and somewhat formulaic responses from the candidate, which closely align with typical AI-generated content focusing on idealized professional scenarios and responses. The conversation flows smoothly with detailed, well-articulated answers that might be overly polished for a typical spontaneous human response in an interview setting." \textbf{Winner: Dual-Prompt Llama 3.3} 
        
        &  

\textbf{GPT-4T interview first}: "[GPT-4o] seems more polished and detailed, with answers that are lengthy and cover multiple aspects of the question, which could indicate AI generation. In contrast, [Llama 3.3] has more concise answers and a more natural flow, which might suggest human involvement. \textbf{Winner: Dual-Prompt Llama 3.3}

\vspace{5pt}
        
\textbf{Llama 3.3 interview first}: "Both conversations seem to be well-structured and coherent, but [GPT-4o] appears more polished and formal, with answers that are slightly more generic and could be generated based on common interview questions and expected responses. [Llama 3.3], while also professional, has a slightly more personal and varied tone, suggesting a human touch." \textbf{Winner: Dual-Prompt Llama 3.3}
        \\        
        \bottomrule
    \end{tabular}
\end{table}

\clearpage
\section{Example Interview Comparison: Single-Prompt GPT-4o vs. Single-Prompt Llama 3.3}

\begin{table}[h]
    \centering
    \small
    \renewcommand{\arraystretch}{1.5} 
    \setlength{\tabcolsep}{10pt} 
    \begin{tabular}{p{0.45\textwidth} p{0.45\textwidth}}  
        \toprule
        \textbf{Two GPT-4o agents in a job interview} & \textbf{Two Llama 3.3 agents in a job interview} \\
        \midrule
        
\textbf{Interviewer}: Thank you for joining us today. To start, can you briefly summarize your professional background for me?

\textbf{Candidate}: Certainly! I have a diverse background, beginning with my role as the owner of a vehicle and heavy machinery business. This allowed me to gain extensive experience in business operations and import-export services. Additionally, I worked in a hotel setting as a front desk representative and an auditor, which helped me hone my customer service and financial oversight skills.

\textbf{Interviewer}: That sounds interesting. Can you tell me more about how you managed the operations in your vehicle and machinery business?

\textbf{Candidate}: As the owner, I was responsible for overseeing the entire operation, which included managing a nationwide dealer network. I utilized auction access to acquire inventory from across the U.S. and coordinated the logistics for import-export services. This required careful planning, relationship management, and negotiation skills to ensure smooth operations and customer satisfaction.

\textbf{Interviewer}: That's impressive. How did your experience in the hospitality industry complement your skills in your business operations?

\textbf{Candidate}: Working as a front desk representative, I developed strong interpersonal communication skills and the ability to address service discrepancies effectively. <truncated>

        & 

\textbf{Interviewer}: Good morning, thank you for coming in today. Can you start by telling me a little bit about yourself and what you're looking for in a professional setting?

\textbf{Candidate}: Good morning, thank you for having me. I'm a highly motivated and versatile individual with a strong background in business operations and customer service. I'm looking for a role that will challenge me and allow me to grow professionally.

\textbf{Interviewer}: That's great to hear. Can you give me an example of a time when you had to handle multiple responsibilities at once? How did you prioritize your tasks and manage your time?

\textbf{Candidate}: Yes, certainly. In one of my previous roles, I was responsible for managing a nationwide dealer network, which involved coordinating with multiple parties and handling various tasks simultaneously. I prioritized my tasks by focusing on the most urgent and important ones first, and then delegated tasks to my team members when possible.

\textbf{Interviewer}: I see. It sounds like you have some experience with leadership and team management. Can you tell me about a time when you had to make a difficult decision that affected a team or an organization? What was the outcome, and what did you learn from the experience?

\textbf{Candidate}: Yes, I recall a situation where I had to make a tough decision to adjust our business strategy in response to changes in the market. <truncated>

        \\

        \midrule
        \textbf{\underline{GPT-4o rationale and preference:}} & \textbf{\underline{Llama 3.3 rationale and preference:}} \\

\textbf{GPT-4o interview first}: "Both conversations exhibit structured and coherent responses typical of AI-generated text, with detailed and articulate answers. However, [GPT-4o] contains more complex and nuanced responses, which might suggest a higher likelihood of AI involvement." \textbf{Winner: Single-Prompt Llama 3.3}

\vspace{5pt}

\textbf{Llama 3.3 interview first}: "Both conversations exhibit structured and coherent responses typical of AI-generated text, with detailed examples and clear articulation of experiences. However, [Llama 3.3] has a more formulaic and comprehensive style, which is often indicative of AI generation." \textbf{Winner: Single-Prompt GPT-4o}
        
        &  
\textbf{GPT-4o interview first}: "Both conversations seem to be well-structured and professional, but [GPT-4o] appears more polished and detailed, with the candidate providing specific examples and skills that align closely with the questions. [Llama 3.3] also provides relevant examples, but the responses seem slightly more generic and less detailed in some areas. However, both conversations could potentially be AI-generated due to their professional tone and structure. After careful consideration, I find both conversations to be equally likely to have AI generation involved due to their similar tone and professionalism." \textbf{Winner: Tie} 

\vspace{5pt}
        
\textbf{Llama 3.3 interview first}: "Both conversations seem to be well-structured and professional, but [Llama 3.3] appears more formulaic and repetitive, with the candidate providing very similar answers to various questions, which could indicate AI generation. In contrast, [GPT-4o ]seems more dynamic, with the candidate providing more varied and detailed responses." \textbf{Winner: Single-Prompt GPT-4o}
        \\      
        \bottomrule
    \end{tabular}
\end{table}

\clearpage
\thispagestyle{empty}

\hypertarget{ReallyLastPage}{}

\begin{center}




\vspace{8mm}

\begin{tcolorbox}[
  enhanced, breakable,
  title=Bib\TeX, fonttitle=\bfseries,
  colback=white, colframe=gray!35,
  boxrule=0.6pt, arc=6pt, width=0.9\textwidth
]
\begin{BibTeXBlock}
@inproceedings{de-baer-etal-2025-single,
    title = "Single- vs. Dual-Prompt Dialogue Generation with {LLM}s for Job Interviews in Human Resources",
    author = {De Baer, Joachim  and
      Do{\u{g}}ru{\"o}z, A. Seza  and
      Demeester, Thomas  and
      Develder, Chris},
    editor = "Arviv, Ofir  and
      Clinciu, Miruna  and
      Dhole, Kaustubh  and
      Dror, Rotem  and
      Gehrmann, Sebastian  and
      Habba, Eliya  and
      Itzhak, Itay  and
      Mille, Simon  and
      Perlitz, Yotam  and
      Santus, Enrico  and
      Sedoc, Jo{\~a}o  and
      Shmueli Scheuer, Michal  and
      Stanovsky, Gabriel  and
      Tafjord, Oyvind",
    booktitle = "Proceedings of the Fourth Workshop on Generation, Evaluation and Metrics (GEM{\texttwosuperior})",
    month = jul,
    year = "2025",
    address = "Vienna, Austria and virtual meeting",
    publisher = "Association for Computational Linguistics",
    url = "https://aclanthology.org/2025.gem-1.74/",
    pages = "947--957",
    ISBN = "979-8-89176-261-9"
}
\end{BibTeXBlock}
\end{tcolorbox}

\end{center}

\end{document}